\documentclass{article}

\usepackage{PRIMEarxiv}

\usepackage[utf8]{inputenc} 
\usepackage[T1]{fontenc}    
\usepackage{hyperref}       
\usepackage{url}            
\usepackage{booktabs}       
\usepackage{amsfonts}       
\usepackage{nicefrac}       
\usepackage{microtype}      
\usepackage{lipsum}
\usepackage{fancyhdr}       
\usepackage{wrapfig}
\usepackage[linesnumbered,ruled,vlined,spanish,onelanguage]{algorithm2e}
\usepackage[ruled,vlined]{algorithm2e}
\usepackage{graphicx}
\usepackage{wrapfig,lipsum,booktabs}

\usepackage{subcaption}
\title{Hyperparameter Tuning for Deep Reinforcement Learning Applications
\thanks{\textit{\underline{Citation}}: 
\textbf{Both authors contributed equally to this research.}} 
}

\author{
  Mariam Kiran*\\
  Scientific Networking Division\\
 Lawrence Berkeley National Laboratory\\USA\\
 \textit{mkiran@lbl.gov}\\
   \And
  Melis Ozyildirim*\\
Department of Computer Science\\Cukurova University\\
Turkey\\
\textit{mozyildirim@cu.edu.tr}\\
 \\
}

\begin{document}
\maketitle

\begin{abstract}
Reinforcement learning (RL) applications, where an agent can simply learn optimal behaviors by interacting with the environment, are quickly gaining tremendous success in a wide variety of applications from controlling simple pendulums to complex data centers.
However, setting the right hyperparameters can have a huge impact on the deployed solution performance and 
reliability in the inference models, produced via RL, used for decision-making. Hyperparameter search itself is a laborious process that requires many iterations and computationally expensive to find the best settings that produce the best neural network architectures. In comparison to other neural network architectures, deep RL has not witnessed much hyperparameter tuning, due to its algorithm complexity and simulation platforms needed. In this paper, we propose a distributed variable-length genetic algorithm framework to systematically tune hyperparameters for various RL applications, improving training time and robustness of the architecture, via evolution. We demonstrate the scalability of our approach on many RL problems (from simple gyms to complex applications) and compared with Bayesian approach. Our results show that with more generations, optimal solutions that require fewer training episodes and are computationally cheap while being more robust for deployment. Our results are imperative to advance deep reinforcement learning controllers for real-world problems.
\end{abstract}


\section{Introduction}
Deep reinforcement learning applications are increasingly being used to teach systems how to drive a car, control massive power grids or self-playing games \cite{RLcontrolSpielberg, alphago}, or finding novel neural network design search \cite{pmlr-v80-liang18b, 199317, 8638041}. With an optimal neural network architecture, a robust deep RL solution can replace controllers in complex environments for model-free optimum control of various complex systems \cite{10.1145/3126908.3126951}. However, a deep understanding of deep RL approaches is required due to the many challenges of discrete versus continuous state representations, exploration variances, and complex reward functions that can learn optimal actions by just observing the environment \cite{8103164}.

Much work on hyperparameter searching, has been extensively studied in applications containing Convolutional Neural networks (CNNs) for image recognition \cite{xiao2020efficient} or general deep learning models \cite{8638041}. This also includes the development of various hyperparameter search libraries such as Ray, Hyperband, DeepHyper, and even hand-tuning to iterate through multiple settings repeatedly until finding fairly tuned hyperparameters such as layer numbers, neurons, and so on. However, deep RL applications, in particular, have comparatively seen less progress in developing methods for optimal hyperparameter search. This is due to many factors - the need for scaling in both the neural network learning and the simulation involved, the multiple RL algorithms developed to handle value-based or policy learning (e.g such as Deep Q-Network (DQN), Deep Deterministic Policy Gradient (DDPG), to name a few \cite{pmlr-v80-liang18b}) or just simple allowing the deep RL to learn via trial-and-error. However, as we aim to replace complex controllers with deep RL solutions, the process of trial-and-error becomes tedious, time-consuming, and computationally expensive, to evaluate and find the best model that is robust. As the problems become more complex, search space grows and resources are limited, there is a need to develop a practical solution targeted specifically for hyper-parameter search in reinforcement learning applications. Within distributed computing research, deep RL research is witnessing an explosion of applications such as controlling data centers for improving the carbon footprint, robotics, optimizing storage demands, or finding optimum traffic engineering solutions to reduce flow completion times \cite{10.1145/3405671.3405809}.

Improving agent actions through multiple interactions was also seen in seminal works by Sim, where creatures used genetic algorithms to evolve limbs to cope with swimming challenges \cite{10.1145/192161.192167}. Without any `learning', evolutionary algorithms are useful for finding solutions to multi-objective optimization problems and proven successful \cite{996017}. Researchers have used genetic algorithms to find optimal settings in a 3-layer CNN \cite{10.1145/2834892.2834896} but argued that they are too slow due to their mutation only nature \cite{xiao2020efficient}. In this paper, we argue that this is not the case and genetic algorithms can be a powerful manner to automate the unconstrained multi-objective search for RL applications. We focus on combining variable-length genetic algorithm search with deep RL problems to develop a scalable approach for finding optimal hyperparameter for deep RL challenges. 

Although many hyperparameter search libraries use RL techniques to find optimal parameters \cite{liang2018rllib, 8638041}, to the best of our knowledge, these do not use evolutionary approaches to find optimal solutions and also are not specifically designed to target deep RL applications. The reasons may be twofold, the assumption of infinite compute cycles for trial-and-error episodes and deep RL research is still in its infancy when it comes to finding real-world solutions using deep RL. Other hyperparameter tuning libraries such as Ray, HyperBand or Oputuna allow hyperparameter tuning, but not for deep RL applications. In deep RL, the neural network needs to train (number of episodes) to learn the best reward situations for state and action pair mappings. Therefore, not only do we need optimal hyperparameters, but also the training period to let the deep RL evolve. The goal is to find the best hyperparameters that can allow the agent to reach optimal reward mappings in fewer episodes. In our work, we develop HPS-RL, designed as a scalable deployment library, that generates a gene population, uses crossover and mutation to quickly arrive at the best multi-objective optimal hyperparameters for each RL experiment. HPS-RL is designed to work with multiple deep RL algorithms, optimization, and loss functions, to find the best model which needs less training time and produces the most robust model that can be deployed in real-world systems. The main contributions of this work include (1) an automated multi-objective search for hyperparameters using genetic algorithms to evolve the best deep RL solutions that train in fewer episodes and produce robust solutions and (2) demonstrate the scalability of our architecture on multiple processors that can leverage multiple thread and parallel processing to improve the overall search time for hyperparameters.

Our work aims to bridge the gap between deep RL solutions and hyperparameter tuning, showing how HPS-RL can significantly impact RL research and applications being designed in both gaming, simulations, and robotics domains, showing further how can be scaled for distributed computations. The project is open-source and can be found (in the supplementary document).

\section{Identifying Hyperparameter Impact in Deep RL}

A reinforcement learning problem is formulated with an agent situated in a partially observable environment, learning from the environment and past data to make current decisions. The agent receives data as environment snapshots with specific relevant features. Depending on the RL algorithm, the agent computes which action to perform depending on the reward values in the current state. The agent then acts to change its environment, subsequently receiving feedback on its action in form of rewards, until the terminal state is reached. The objective is to maximize the cumulative reward overall actions in the time agent is active \cite{Sutton:2018:RLI:3312046}. Deep RL research has released a set of environments (or gyms) to develop new deep RL algorithms that can learn optimal actions faster. OpenAI gym experiments are a collection of many control, gaming, and robot problems that users can use to train their deep RL algorithm against \cite{deepq}.

\textbf{Hyperparameters Impact.} As seen in \cite{Henderson_dlmatters}, the sensitivity of various deep RL techniques can impact the reward scale, environment dynamics, and reproducibility for the experiments tried. Finding benchmarks for a fair comparison is a challenge, but OpenAI stable baselines can help provide initial comparison across the gym environments chosen. Complex hyperparameters can impact how quickly the RL solutions reach consensus or produce a robust solution that has enough experience to cope in the environment and achieve the maximum reward. Mostly grid search is used to find optimal parameters \cite{pmlr-v80-liang18b}.
\vspace{-1mm}
\begin{equation}\label{eq:rl}
    Q(S_{t}, A_{t})\leftarrow Q(S_{t}, A_{t}) + \alpha (R_{t+1}+ \gamma Q(S_{t+1}, A_{t+1})-Q(S_{t}, A_{t}))
\end{equation}

Equation \ref{eq:rl} describes the Bellman equation learning optimal action-state pairs from previous states. Exploring the \emph{learning rate} ($\alpha$) can allow the agent to learn quickly in fewer trial episodes, but not necessarily exploring the space of possibilities. \emph{Gamma} ($\gamma$) helps the agent to prioritize immediate rewards versus future rewards in building optimum action strategies. Additionally, agents can use a probability \emph{epsilon} ($\epsilon$) to prioritize greedy action searches taken to learn more robust solutions.

\begin{figure}
\centering
  \begin{subfigure}
    [b]{0.4\textwidth}
    \includegraphics[width=0.95\linewidth]{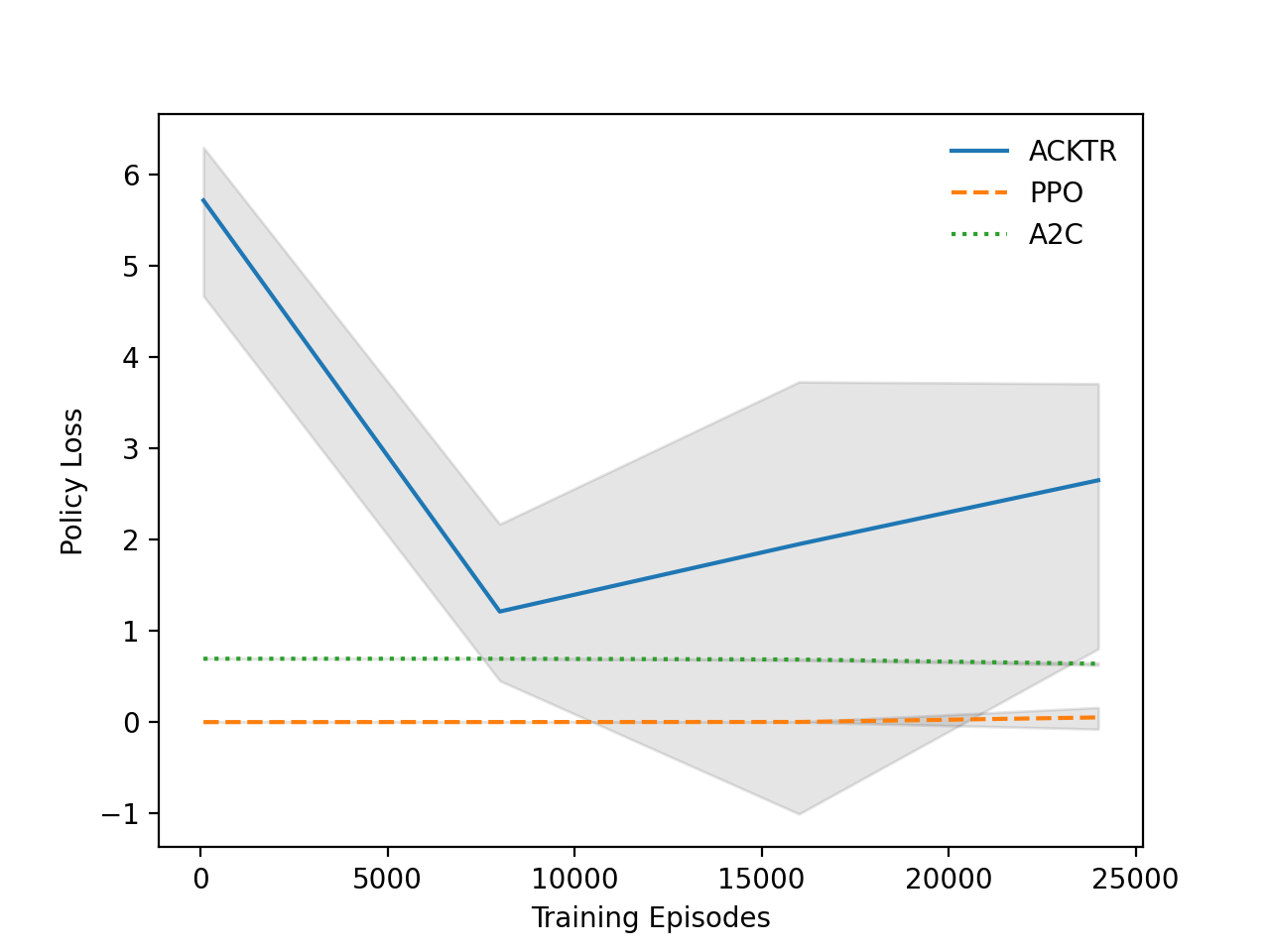}  \caption{Policy loss.}
    \label{fig:Tsnejlab}
\end{subfigure}
\begin{subfigure}
   [b]{0.4\textwidth}
\includegraphics[width=0.95\textwidth]{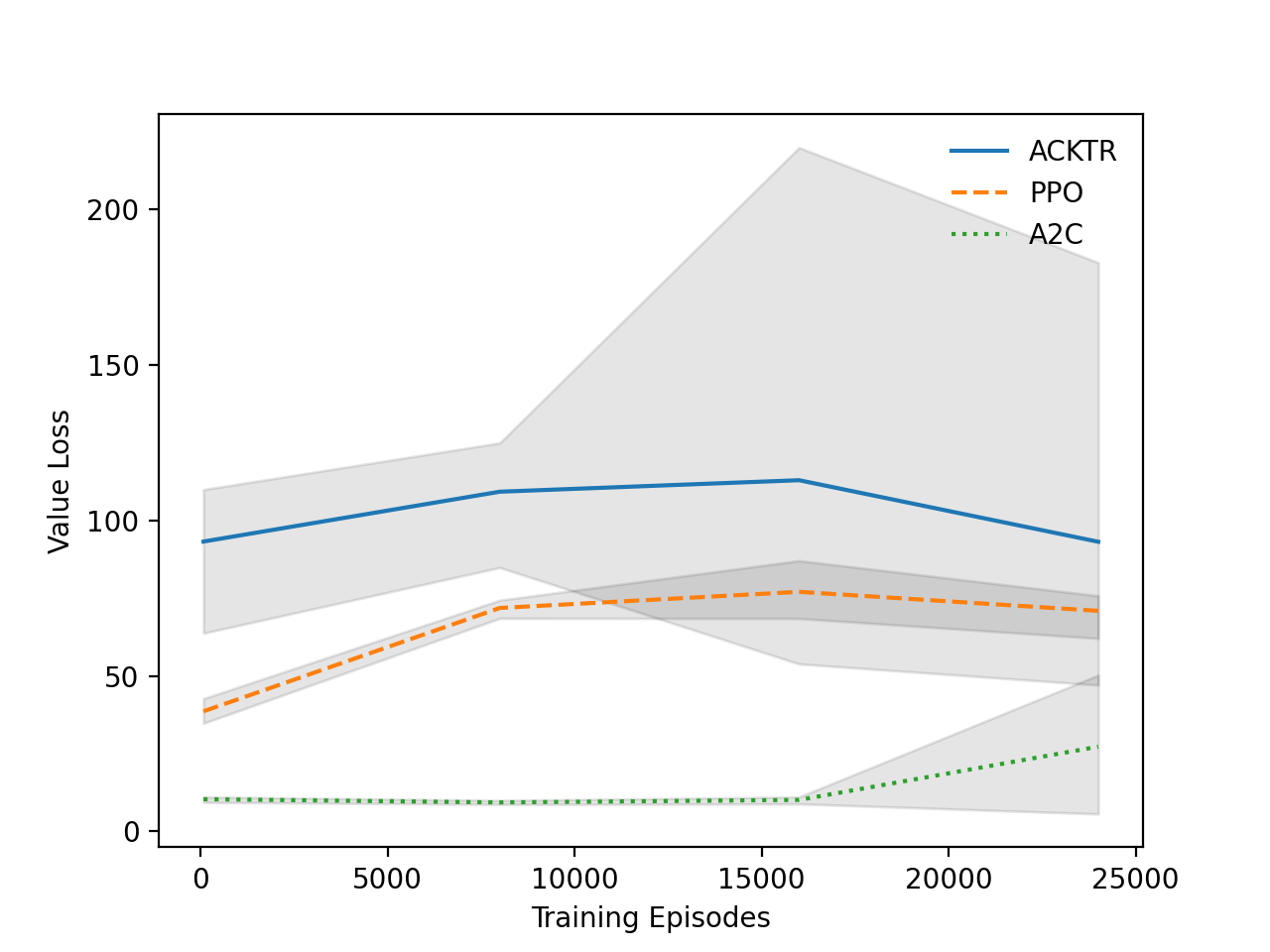}
    \caption{Value loss.}
    \label{fig:Tsnepnnl}
\end{subfigure}
 \caption{Baseline runs in Cartpole Environment showing impact of different RL algorithms on the same problem.}\label{fig:tsneuser}
  \label{fig:cart-baseline}
\end{figure}

\textbf{Deep RL Algorithms.} Most deep RL problems can be framed as Markov Decision Processes (MDPs), consisting of four key elements $\langle \mathcal{S}, \mathcal{A}, P$, $R\rangle$, taking decisions at each epoch or episode $t$. The probability that the agent moves into new state $s_{t+1}$ is given by the state transition function $P(s_{t+1}|s_t,a_t)$. In model-free RL, the agent aims to learn a policy to learn how to act. Here, \textbf{policy gradient functions} $\pi(a_t|s_t)$ learn a policy mapping from state $s_t$ to action $a_t$ and tells the agent which action $a_t$ to take in-state $s_t$. \textbf{Value function} $V_{\pi}(s_t)/Q_{\pi}(s_t,a_t)$ aim to modify the agent's policy based on perceived value of state. Q-learning, deep Q-learning is value-based algorithms, whereas actor-critic, DDPG, leverage both value and policy-based learning to reach a better robust trained agent. Additionally, other examples of algorithms such as TRPO (trust region policy optimization), or PPO (proximal policy optimization), or more, are designed to allow the agent to learn optimal policies quicker, again improving training time. In deep RL literature, most of these algorithms have been developed to tackle discrete versus continuous state and action spaces or improve the training of the agent without compromising the robustness of the solution produced \cite{henderson2019deep}. Figure \ref{fig:cart-baseline} shows the variability in the rewards learned in multiple deep RL algorithms on the same Cartpole Gym environment from OpenAI. This shows it is important to try out multiple algorithms that is best suited for the application being designed. Various research has developed various versions of deep RL techniques that range from studying and acting in discrete versus continuous states \cite{mnih2013playing}, improving the exploration space or the memory buffers which allow the agent to learn from past actions. Combined with variance in methods this poses challenges in reproducibility, where various techniques have shown variable results against baselines, creating a need to investigate how hyperparameters can help build reliable deep RL solutions \cite{Henderson_dlmatters}. Selecting the right algorithms, either value-based or policy-based, can have a significant impact on the results produced \cite{islam2017reproducibility}. 

\textbf{Environment Gyms.} Gym (\url{https://openai.com/}) is a library that provides various reinforcement learning environments to develop and compare algorithms. It is frequently used as a standard environment with several categories such as Atari games, classical control problems, robotics, etc.

\vspace{-2.5mm}

\section{Bayesian Optimization for Hyperparameter Search}

Bayesian optimization methods emerged as an efficient way of hyperparameter search for the problems when the convexity is not guaranteed. When there is no information for fitness function, approximate function is obtained by using Gaussian Kernels which are one type of surrogate models. The next step is finding a function $H$ which shows the next best point. There are four types of $H$ function, which are generally preferred; expected maximization, maximum probability of maximization, and upper confidence bound, entropy search. Although finding the best point by using $H$ is another optimization function itself, gradient based optimization is applied. Algorithm \ref{alg:bohps} shows the Bayesian Optimization process \cite{BayesianGeneralInfo,BayesianGeneralInfo2}.

\begin{algorithm} [H]
\caption{Bayesian Optimization Using Gaussian Kernels}
\SetAlgoLined
\label{alg:bohps}
Evaluate fitness function $f(x)$ on $n$ random points \\
\For {n = 1 to N}{ 
Update Gaussian Posteriors by using all data in $(X_{1:n},f(X_{1:n}))$ \\
Compute next best point function $H(x,X_{1:n})$ using the Gaussian Kernel \\
Evaluate $f(x)$ on $x_n+1$ which maximizes the $H(x,X_{1:n})$ \\
Add $x_n+1$ to $(X_{1:(n+1),f(X_{1:(n+1)}))}$ \\
} 
Return $x_i$ evaluated with the largest $f(x_i)$
\end{algorithm}

\section{HPS-RL: Hyperparameter Search using Genetic Algorithm (GA)}
The design of HPS-RL is a multi-objective optimization problem. In this paper, we use a genetic algorithm (GA) based search method to perform this multi-objective optimization to allow a trade-off between competing variables in the hyperparameter space. 

Genetic algorithms have been used as optimization methods in multiple research areas \cite{goldbergga}. Modeling the search routines as a Darwinian evolution theory allows optimum (or fittest) solutions to emerge from a group or population of solutions. Experiments in \cite{10.1145/192161.192167} show that evolution was a successful strategy to produce solutions that survive the longest or, in our setting, more robust for RL challenges. For HPS-RL the GA requires:
\vspace{-1.0mm}
\begin{itemize}
    \item Developing a scheme to represent the hyperparameters as a gene, which is part of a population (of many individuals).
    \item Developing a `survival of the fittest' method, to measure which gene performs well - the selection process.
    \item A computationally efficient method to evaluate the individuals for trials - highest rewards achieved in fewer training episodes. 
    \item A method to perform crossover and mutation, to generate new individuals from past well-performing individuals, for a new population generation.  
\end{itemize}

\begin{figure*}
  \centering
    \includegraphics[width=0.85\linewidth]{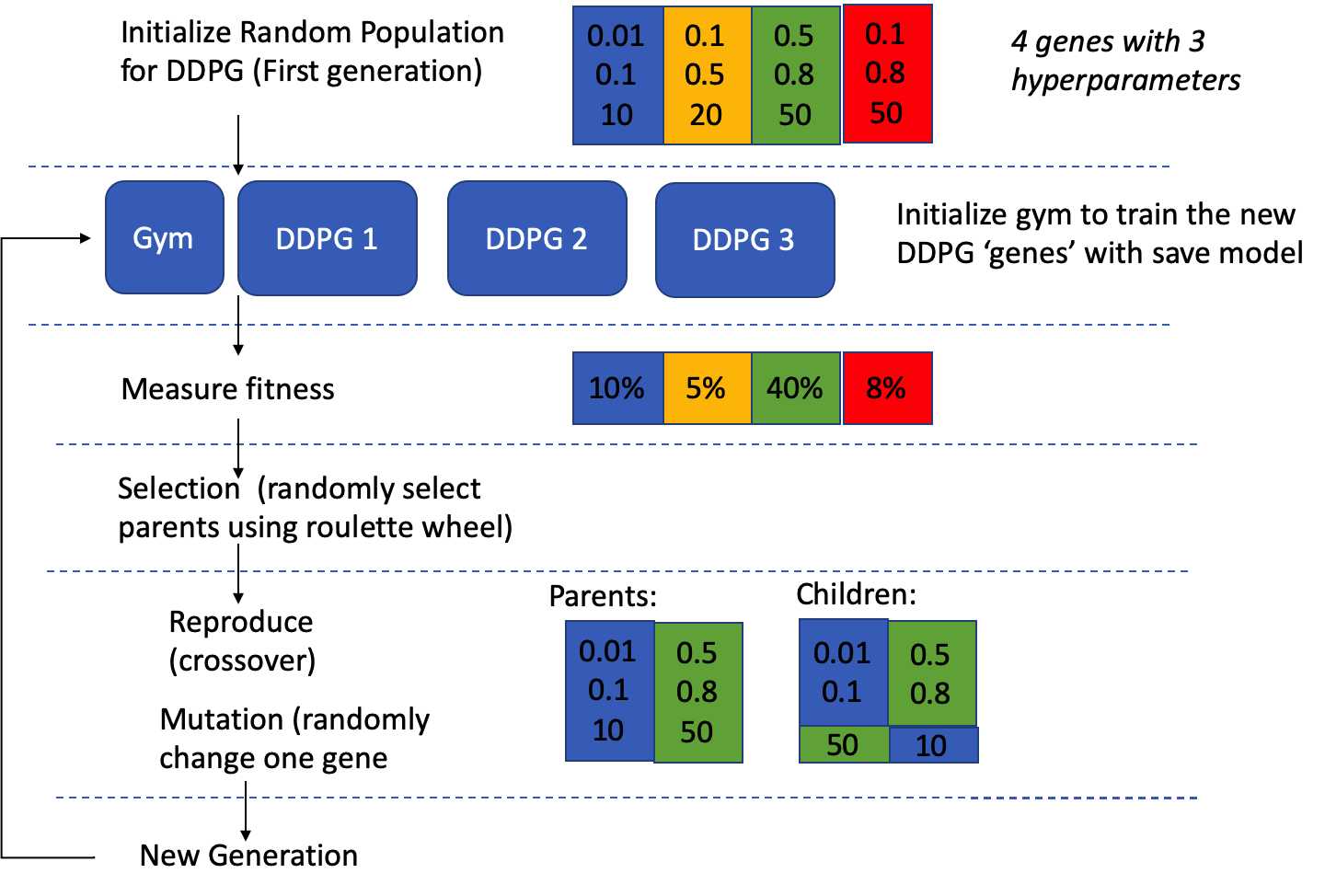}  \caption{Example GA search for the best hyperparameters for DDPG in a chosen Gym.}
  \label{fig:gageneral}
\end{figure*}

\subsection{Selection and Fitness Mechanism}
Figure \ref{fig:gageneral} shows the overall process of HPS-RL. We initialize individuals with a block of hyperparameters being explored. For example, in a DDPG case, the basic gene block looks like $(\gamma, \alpha, no. of neurons)$. Additional hyperparameters such as activation functions, $\epsilon$, number of layers, and more, can also be explored. Based on the example, with 4 blocks of individuals, each set is trained with the chosen gym environment and the model is saved. Then the saved model is evaluated for fitness by playing with the gym and recording the cumulative reward playing for 100 steps.

After fitness evaluation, the roulette wheel selection mechanism can allow a random selection among the well-performing parents \cite{backThomas},
$ p_{i} = \frac{f_{i}}{\sum _{j=1}^{N} f_{j}}
$, where $N$ is the number of individuals in the population and $p_{i}$ allows a higher probability of the `fitter' parents to be chosen for the next generation. The fitness of the parents is calculated using $    f_{i} = \frac{1}{n} +\sum_{k=1}^{n} r_{k} + \frac{1}{\sum_{k=1}^{n} l_{k}}$, where $n$ is the number of episodes being played, $r$ is cumulative over $n$ episodes and $l$ represents the total loss incurred during this evaluation.

\textbf{Crossover and Mutation.} The new generation of selected individuals become parents of the new offspring by performing crossover and mutation properties. As shown in Figure \ref{fig:gageneral}, at a random point genes are swapped from one parent to the second, to generate two new individuals. For example, represented ($\gamma, \alpha, no. of neuron$) - Parent 1 (0.01,0.1,10) and Parent 2 (0.5,0.8,50), will become - Child 1 (0.01,0.1,50) and Child 2 (0.5,0.8,10). 

In mutation, we select one variable and randomly replace it. For example Child 1(0.01, 0.1, 50) will become - Child 3 (0.01,\textbf{0.5},50). The new generations are then evaluated again via training and fitness measurements to allow better generations (or solutions) to be obtained. The crossover and mutation rates can be altered to either allow a slower change in population dynamics or to allow the system to explore multiple optimal fronts to find the best solutions. 

\textbf{Variable-length genes for the algorithms.} Deep RL research has developed many deep RL algorithms ranging from DQN, DDQN, DDPG to complex algorithms ACKTR. Each algorithm differs by the optimization functions and manners in which the rewards are learned. With this challenge, the hyperparameters that can be tuned in each algorithm are based on the optimization functions as part of their architectures. For example, a DDPG gene ($\alpha, no. layers, neurons$) is different from an ACKTR gene ($\alpha, no. layers, neurons, KL-divergence$).
Algorithm \ref{alg:gahps} shows how GA can help find optimal solutions.

\begin{algorithm} [H]
\caption{GA for HPS-RL}
\SetAlgoLined
\label{alg:gahps}
Initialize gene population $p$, rate of crossover $c$, rate of mutation $m$, evaluation episodes $e$\\
Generate $p$ possible hyperparameter solutions to explore
\\
\For {episode = 1 to p}{ 
Train the deep RL chosen with each solution\\
Save the respective deep RL models\\
Evaluate the trained models, by evaluating the saved models for $e$ episodes\\
Determine the fitness of the solution evaluated, reward, training episodes and loss incurred\\
Perform roulette wheel selection\\
Select parents\\
Perform crossover and mutation depending on rates ($c,m)$\\
Generate children and add to new population generation to try $p+1$\\
}
Return best solutions in $p$
\end{algorithm}

\section{Software Description}
\begin{figure}
  \centering
    \includegraphics[width=0.6\linewidth]{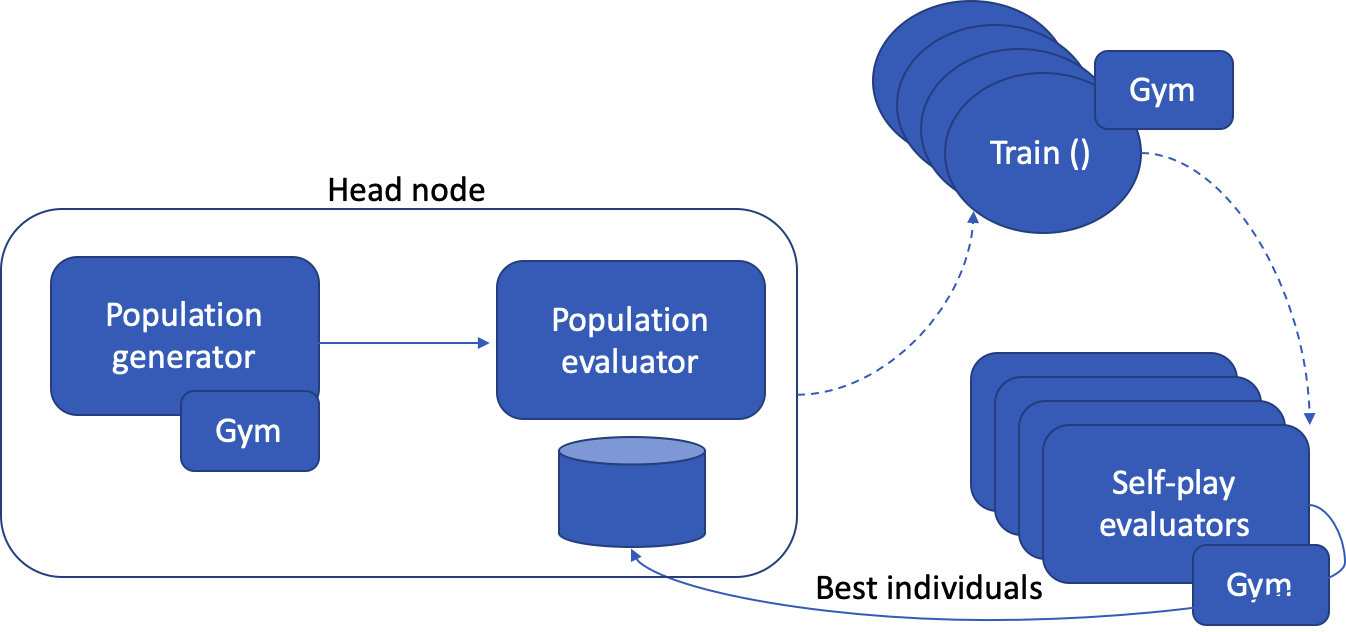}  
    \caption{Distributed HPS-RL architecture. The head node can set up multiple nodes to run multiple training and evaluation nodes for each individual. All workers will update the parameter server in the head node to find the optimal hyperparameter for the deep RL application.}
  \label{fig:gadist}
\end{figure}

Our open-source HPS-RL software contains three Python subpackages: (1) a collection of genetic algorithm functions, (2) benchmark gym environments, (3) benchmark deep RL algorithms with optimization functions, to explore. The current suite can be extended with mpi4py implementations for distributed computations. Currently, we run these models as multiple collections of jobs from a single head node. Figure \ref{fig:gadist} shows the high-level architecture of how the head node generates the population and maintains a parameter server that collects the fitness among all the individual searches. 
HPS-RL is designed as a distributed library and can be extended on HPC platforms. Currently, we exploit the multiprocessors. The hardware we used for our testing has Intel(R) Core(TM) i7-9750H CPU @ 2.60GHz, 64 GB RAM, 12 cores, and Nvidia GeForce RTX 2070 GPU processor which has 2304 Cuda cores and 8 GB RAM. 

\subsection{HPS-RL Deep RL algorithms}
HPS-RL packages its deep RL algorithms for researchers to explore. HPS-RL provides more deep RL algorithms listed in Table \ref{tab:tunedparams}.

\subsection{HPS-RL Optimization functions}
HPS-RL packages its optimization function variants, to allow researchers to explore more approaches, other than the standards stochastic gradient descent \cite{ozyildirim2020optimization}. Gradient descent is a common optimization algorithm for training neural networks. Based on the idea of updating the tunable parameters to minimize the objective function, and uses the learning rate to converge the loss function. Conjugate gradient works on finding an optimal coefficient $\beta$ to prevent fluctuation. CG leads to a fast convergence, it often results in poor performance. Approaches like Broyden-Fletcher-Golfarb-Shanno Algorithm use Quasi-Newton methods approximate the Hessian value to solve unconstrained optimization problems. BFGS algorithm is one of the efficient Quasi-Newton approaches. Levenberg - Marquardt Algorithm is an optimization method for solving the sum of squares of non-linear functions. It is a combination of gradient descent and the Gauss-Newton method. It starts with gradient descent and as it comes close to the solution, it behaves like a Gauss-Newton method, which it achieves by using a damping factor. While a larger damping factor results in gradient descent, a small damping factor leads to the Gauss-Newton method.

\section{Experimental Results}
In this section, we evaluate HPS-RL and its capability to find optimal parameters for deep RL applications. We evaluate the experiment on three deep RL applications (gym environments).
\subsection{Gym Environments}

\begin{table}
\centering
\begin{tabular}{lllll}
\hline\noalign{\smallskip}
 Parameter &  
DQN & DDPG & TRPO & A2C
 \\
\noalign{\smallskip}\hline\noalign{\smallskip}
Episodes\\
(50, 200, 500) & $\checkmark$   & $\checkmark$  & $\checkmark$  & $\checkmark$\\
\hline
Gamma  $\gamma$\\
(0.01,0.1,0.5,0.99)& $\checkmark$   & $\checkmark$  & $\checkmark$  & $\checkmark$\\
\hline
Learning rate\\
(0.1, 0.01, 0.001)& $\checkmark$   & $\checkmark$  & $\checkmark$  &  $\checkmark$\\
\hline
Batch Size& $\checkmark$   & $\checkmark$  & $\checkmark$  &  $\checkmark$\\
\hline
Neurons No.& $\checkmark$   & $\checkmark$  & $\checkmark$  &  $\checkmark$\\
\hline
Layers & $\checkmark$   & $\checkmark$  & $\checkmark$  &  $\checkmark$\\
\hline
Optimization function \\(Adam, CG LBFGS, LM)& $\checkmark$   & $\checkmark$  & $\checkmark$  &  $\checkmark$\\
\hline Activation function \\(tanh, relu)& $\checkmark$   & $\checkmark$  & $\checkmark$  &  $\checkmark$\\
\hline
Trajectory Size \\(10, 20, 50, 100, 1000)&   &   &  & $\checkmark$ \\
\hline
KL value \\(0.001, 0.01, 0.1)&   &   &  &  $\checkmark$\\
\hline
\noalign{\smallskip}\hline
\end{tabular}\caption{Hyperparameters for deep RL algorithms.}\label{tab:tunedparams}
\end{table}
\textbf{\emph{Cartpole Environment.}} Part of the OpenAI Gym environments, CartPole is a classical control problem of an inverted pendulum problem which is connected to a pole, to balance the pole for as long as possible \cite{gymdataset}. The pole is linked to a cart, which moves along a track. The state, action details are:
\textbf{State:} The agent reads in 4 continuous variables which represent the current environment, position of the pole, cart, angle, and velocity.
\textbf{Action:} The action is continuous to move left or right by a certain amount of degrees. 
\textbf{Reward:} A reward of +1 is obtained every time step the pole remains upright. The episode ends when the pole is more than 15 degrees from vertical.

\textbf{\emph{Lunar landing.}} Another environment from OpenAI gym, this game involves trying to control the landing of a ship on a landing pad. The state, action, and reward are as follows: 
\textbf{State:} The state observation space is the current image of the environment represented as 8-dimension continuous variables.
\textbf{Action:} The action space is discrete, \emph{do nothing, move left,} or \emph{move right}.
\textbf{Reward:} The agent receives a reward every time it moves if it crashes the episode ends and the game restarts. Additionally, there is +100 or -100 depending on the success of the landing.

\textbf{\emph{Autonomous laser control.}} Taken from a real experiment, that is using deep RL to control high power lasers for advanced particle accelerators, this experiment aims to perform coherent beam combining (CBC) \cite{Mohammed:20}. The state, action is as follows:
\textbf{State:} The state observation space is a 5x5 matrix of continuous variables representing the intensity of the refracted beams. 
\textbf{Action:} The action is a 3x3 continuous space, tuning 9 beams to achieve maximum combining efficiency.
\textbf{Reward:} Every time the beam moves it receives a +1 reward and a major +100 when the combining has achieved maximum values.

\subsection{Baseline Comparisons}

With many RL algorithms, OpenAI gym also released stable baseline implementations of many algorithms DQN, DDPG, TRPO, A2C, and more. Working on discrete and continuous observation and action spaces, the algorithms are aimed to allow the research community to test algorithms quickly to see the impact of the deep RL in the environment being built \cite{stable-baselines}. Figure \ref{fig:cart-baseline} are stable baseline outputs for Cartpole, showing the variance among the different deep RL algorithms chosen for the same problem. These graphs show why it is important to have a hyperparameter search library that can find the best hyperparameters for the deep RL application.

As deep RL research matures, having more control of hyperparameters and internal functions is needed to develop new deep RL algorithms. HPS-RL is released with its library of deep RL algorithms that allow researchers more flexibility to try multiple function enhancements, potential leading to new RL algorithms. These implementations are based on \cite{lapanbook}. Table \ref{tab:tunedparams} shows the hyperparameters that can be controlled via GA in HPS-RL in these algorithms. ACKTR and A2C use formulas that allow KL convergence to find trust regions to explore and find optimal values. These formulas are not part of other deep RL algorithms and thus not tuning parameters for those.

\begin{figure*}
\centering
  \begin{subfigure}
  [b]{0.3\textwidth}
    \includegraphics[width=1\linewidth]{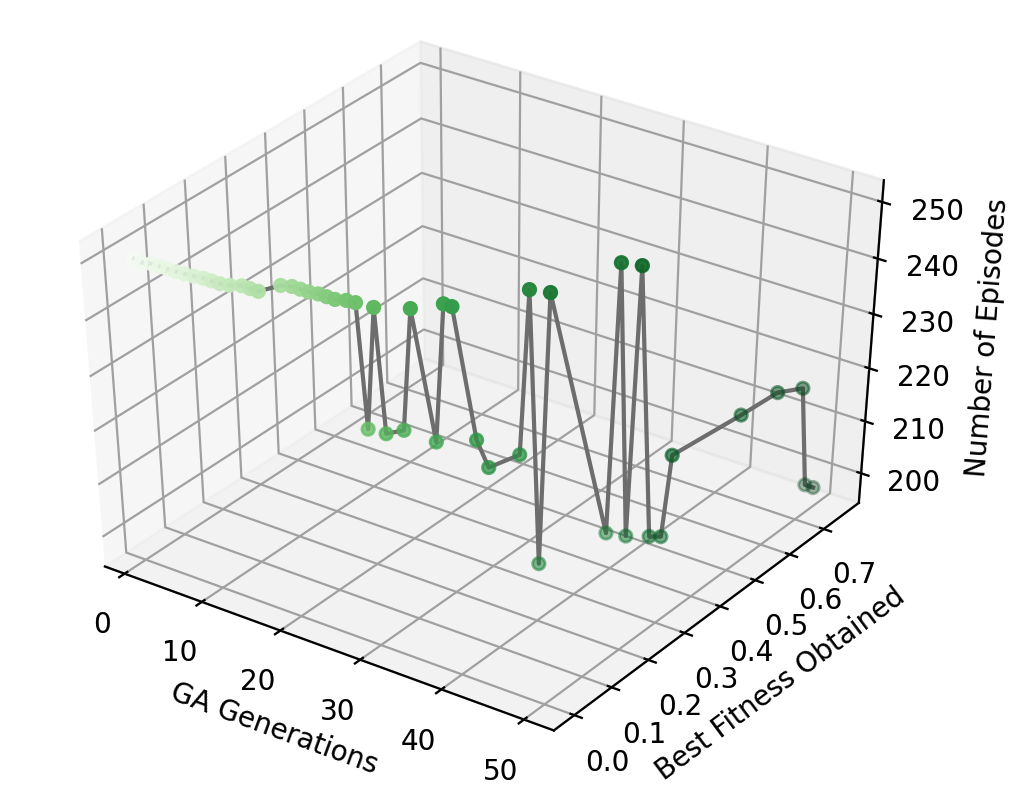}  \caption{Cartpole.}
    \label{fig:cartpolefitness}
\end{subfigure}
\begin{subfigure}
[b]{0.3\textwidth}
\includegraphics[width=0.95\textwidth]{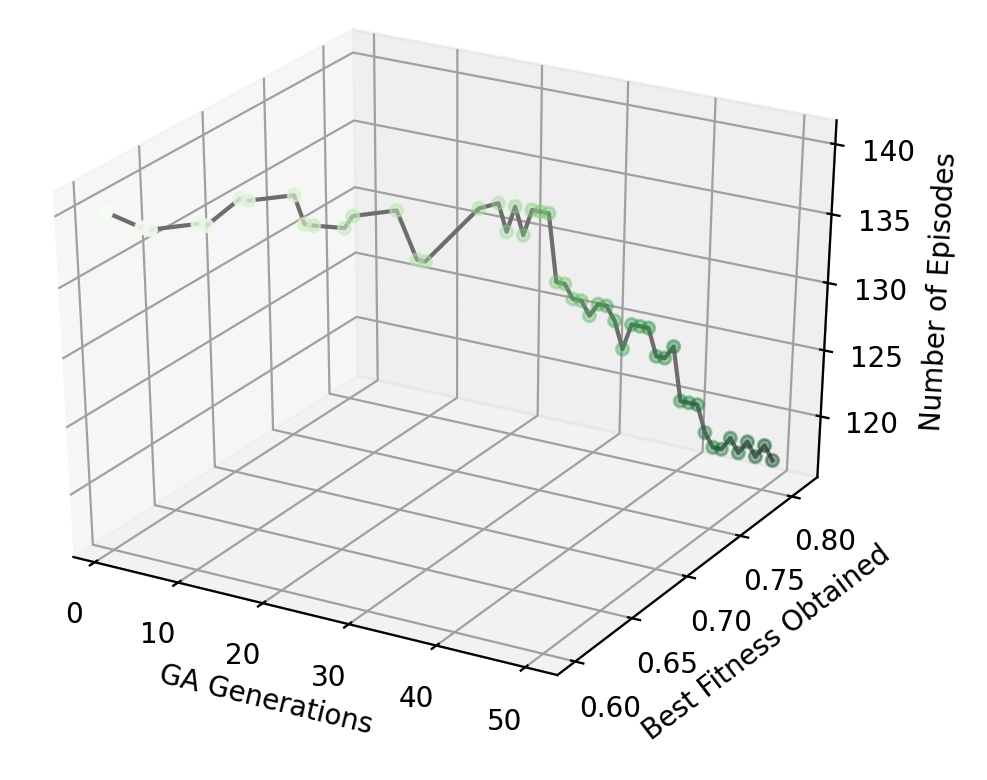}
    \caption{Lunar Landing.}
    \label{fig:lunarfitness}
\end{subfigure}
\begin{subfigure}
[b]{0.3\textwidth}
\includegraphics[width=0.95\textwidth]{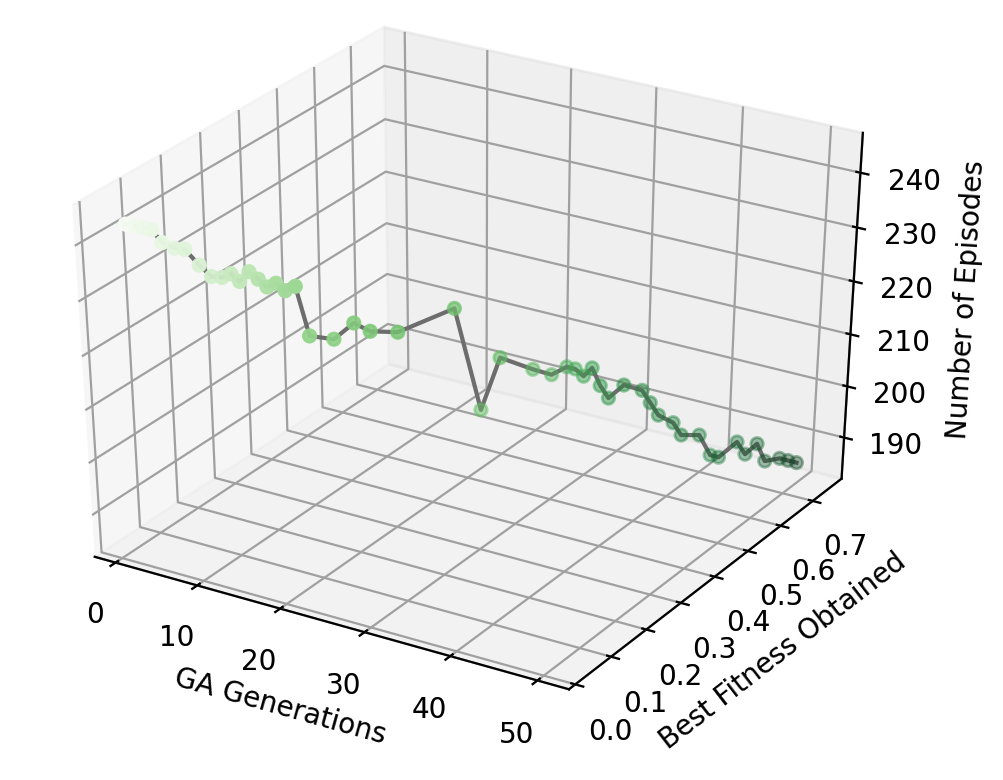}
    \caption{Autonomous Laser.}
    \label{fig:laserfitness}
\end{subfigure}
 \caption{More generations find optimal deep RL solutions, better fitness and less training episodes.}\label{fig:bestfitnessr}
\end{figure*}

\subsection{Evaluation of GA search}
In this section, we describe the results of the search for the best hyperparameters for the 3 gym environments chosen. We evaluate the results to find how many generations does it take to find the best fitness and eventually best performing deep RL neural network architectures.

\begin{wraptable}{r}{5.5cm}
\centering
\begin{tabular}{lll}
\hline\noalign{\smallskip}
Generation & LR & Gamma\\
\hline
GA0 & 0.001 & 0.025\\
\hline
GA1 & 0.01 & 0.025\\
\hline
GA2 & 0.1 & 0.025\\
\hline
GA3 & 0.001 & 0.25\\
\hline
GA4 & 0.01 & 0.25\\
\hline
GA5 & 0.01 & 0.25\\
\hline
\noalign{\smallskip}\hline
\end{tabular}\caption{Best performing individual in each generation.}\label{table:structures}
\end{wraptable}

\begin{figure*}
  \centering
    \begin{subfigure}
[b]{0.45\textwidth}
\includegraphics[width=1\linewidth]{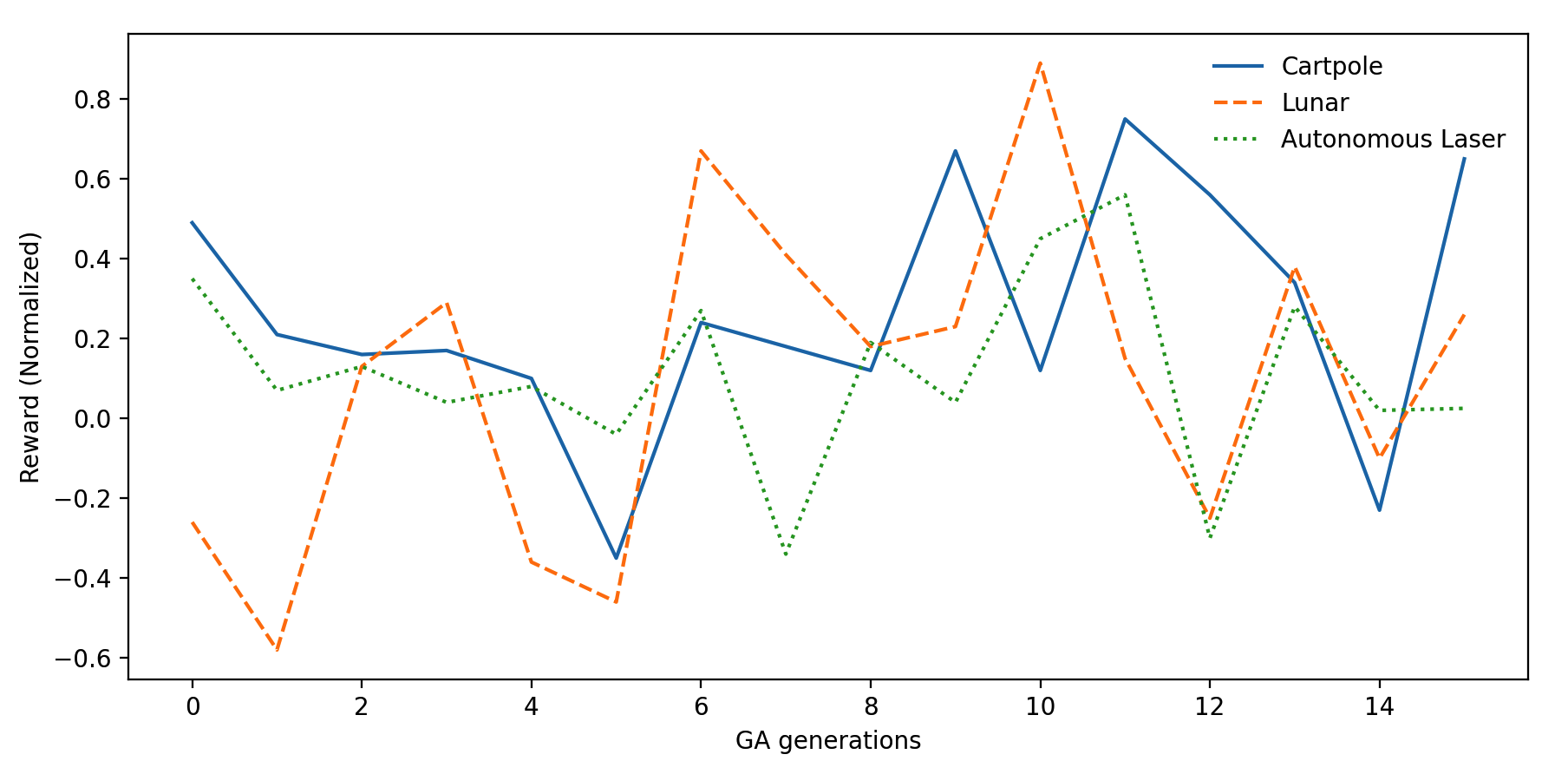}  
    \caption{Rewards show an upward trend as new generations are produced.}\label{fig:reward}
\end{subfigure}
 \begin{subfigure}
[b]{0.45\textwidth}
 \includegraphics[width=1\linewidth]{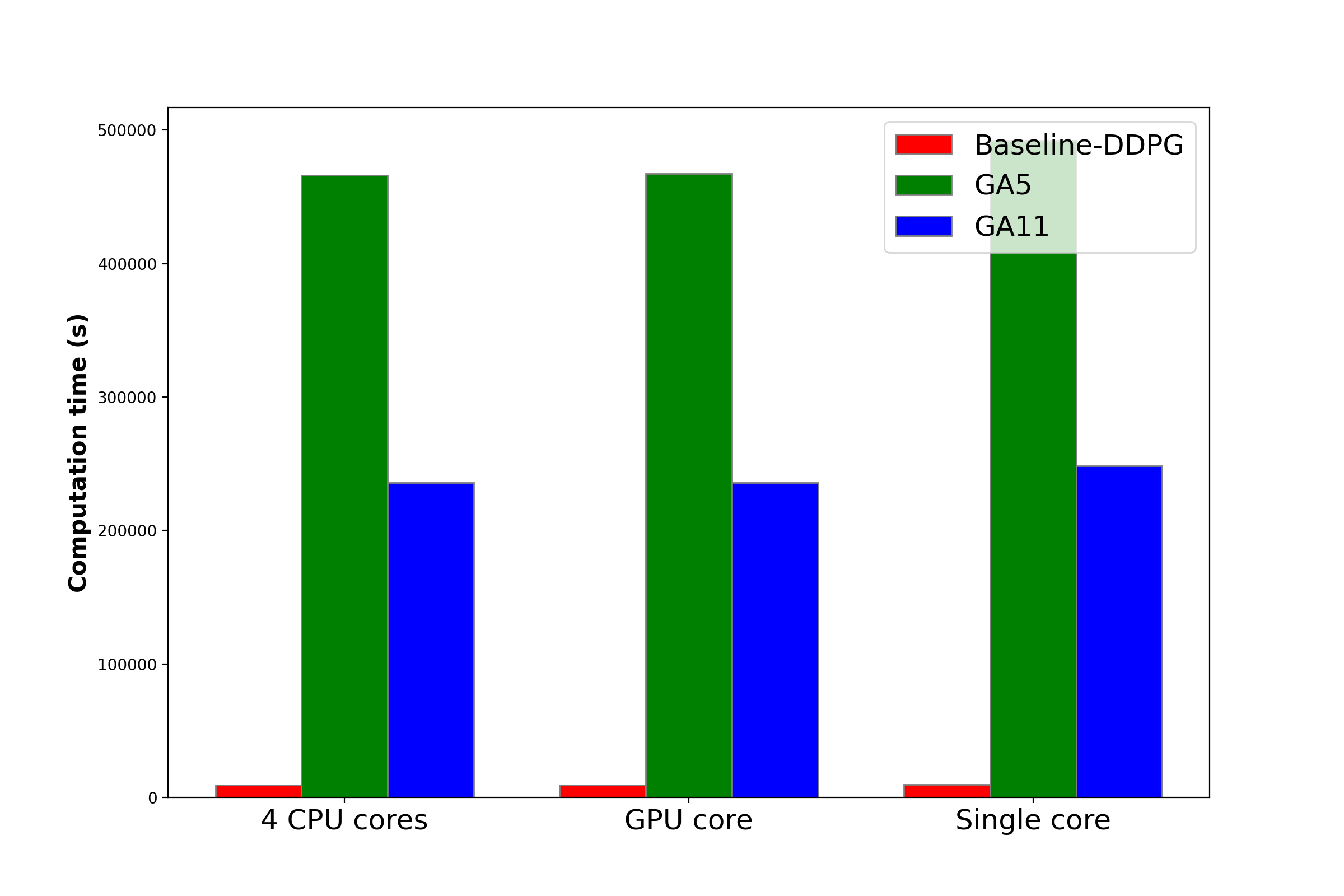}  
    \caption{Computational time.}\label{fig:time}
\end{subfigure}
 \caption{HPS-RL results.}\label{fig:resultsbestfitnessr}
\end{figure*}

Figure \ref{fig:bestfitnessr} shows the comparison of generations, with the number of episodes and fitness achieved. We wee that in all gym environments, the more generations (up to 50) the solutions found can converge to better fitness and also less number of episodes required for training the algorithms. This shows the solutions have evolved to compute hyperparameters that can build robust deep RL solutions, that take fewer episodes for trial-and-error to learn about the system. These produced solutions are more robust for the problems. Figure \ref{fig:reward}
shows different GA structures and normalized reward values for the tested gym environments. We ran the experiment for 44 generations, and best individuals in each generation was captured in Table \ref{table:structures}.

\subsection{Comparison of Multi-core Run Time}
The proposed approach is compatible with GPU, multicore CPUs and single core implementations. During the tests, 4 of 12 CPU cores and all CUDA cores were utilized and Python concurrency library is used for parallelism. The graph shows Cartpole baseline implementation, the tested  GA structure GA-5, and  GA structure GA-11. The baseline implementation does not include any HPS-RL to find optimal parameters, it is just one run and the fastest approach. However, when we start looking for optimal hyperparameters using HPS-RL, GA-5 shows that this takes more time than GA-11 which requires a fewer number of episodes. Moreover, these comparisons show that running on GPU or 4 CPUs performs similar, but increases the computational time of GPU. 
With extensions using mpipy, this processing time can be further improved with parallelism, to be extended in future.
\section{Related Work}

\textbf{RL Applications:} Reinforcement learning applications are being studied for investigating self-learning or self-optimizing behaviors in complex decision-making scenarios, e.g. self-driving cars \cite{henderson2019deep}. The authors also discuss the impact hyperparameters can have on the baseline performance of an RL solution. These include neural network architecture, reward scale, and random seeds. Raising interesting concerns on reproducibility, making an important case for benchmark RL solution as a generalizable solution. In \cite{8125811}, different RL algorithms such as value-function approximation, actor-critic policy gradient, and temporal-difference were compared on the Cartpole benchmark system. The authors also proposed a method for integrating RL and swing-up controllers. 

\textbf{Hyperparameter search:} RL itself can be used to find optimum parameters such as in CIFAR and other complex CNN applications \cite{zoph2017neural}. However, these searches often assume infinite compute resources to find optimum solutions over millions of hours, which is always not the case. 

Similar to HPS-RL, population-based searches have been used to tune hyperparameters using a population of agents, and shown to be effective for RL applications. The authors also argue about the computational limitations to find optimum solutions in small research labs
\cite{parkerholder2021provably}. Wang et al \cite{wang2020loss} tuned loss functions to optimize the search space based on reward versus random. This was also seen in \cite{jomaa2019hyprl}, where they tune validation loss to yield good models in fewer training episodes. Further, researchers \cite{pmlr-v80-falkner18a} combined Bayesian optimizations with bandit solutions to produce solutions better than available hyperparameter tuning libraries such as HyperBand. 

\textbf{Hyperparameter tuning Libraries.} There are many libraries defined for hyperparameter tuning. Each of them has different strengths, compatibility, and weakness. Some of the frequently used libraries are Ray-Tune which supports state of the art algorithms such as population-based training, Bayes Optimization Search (bayesOpSearch); Optuna which provides easy parallelization; Hyperopt which works both serial and parallel ways and supports discrete, continuous, conditional dimensions; ml machine utilizing Bayesian optimization; Polyaxon which is for large scale applications \cite{parameterlib}. In this work, however, HPS-RL is specifically designed to target deep RL applications to find optimal and robust deep RL solutions for control or gym environment problems, which is, to be best of our knowledge, still missing in current implementations.

\subsection{Comparison with Bayesian Optimization}
\begin{figure}
  \centering
\includegraphics[width=0.7\linewidth]{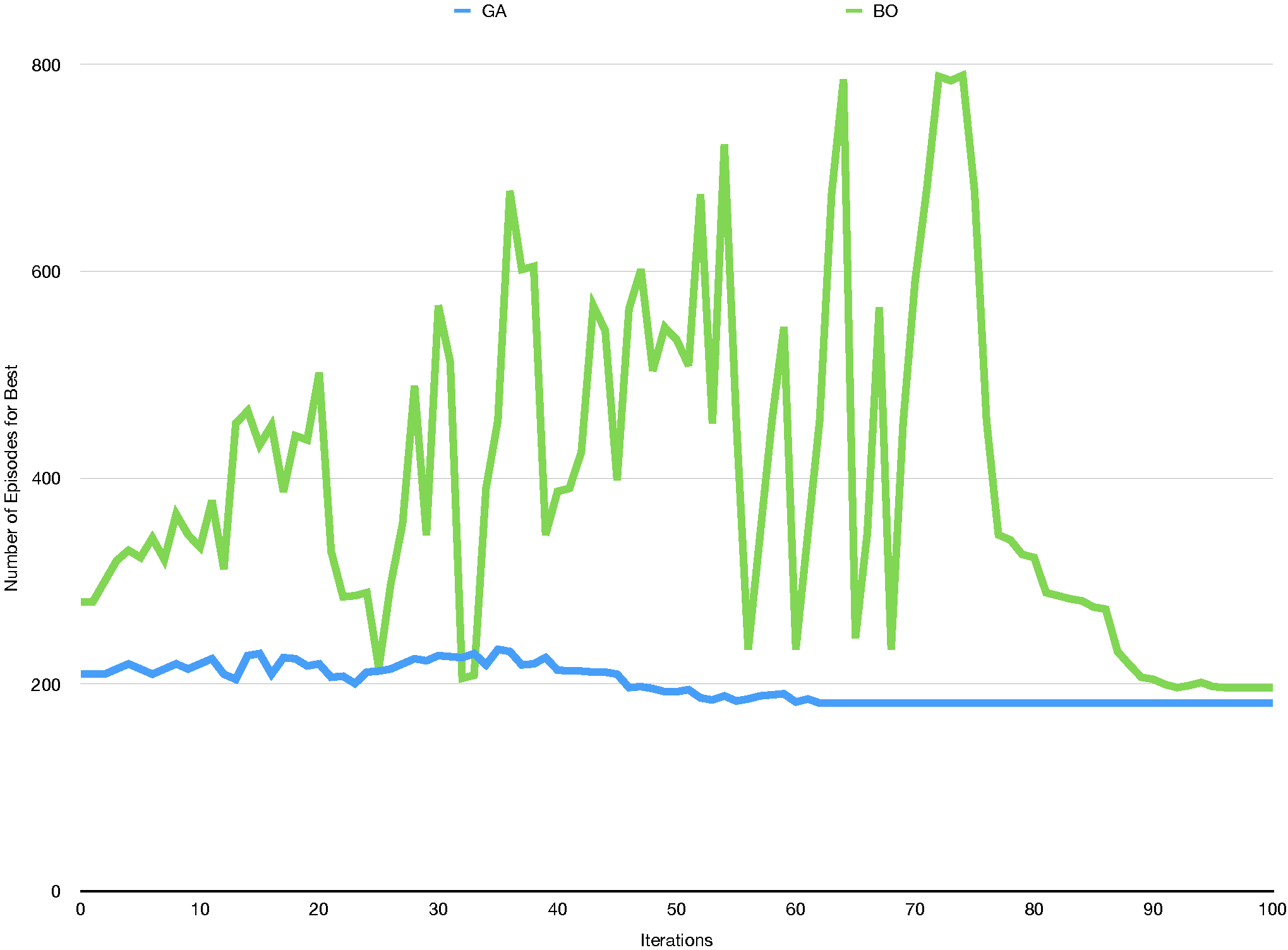}  
    \caption{Number of Episode comparison of GA and BO for achieving best value.}\label{fig:BO}
\end{figure}

The larger number of hyperparameters to be searched, the lower efficiency is obtained from Bayesian optimization \cite{BayesianOptComp}. In this section, we compared the number of iterations required to reach the optimal solutions with GA and Bayesian Optimization approach on Cartpole env. Figure \ref{fig:BO} shows the comparison.

\section{Conclusions}

We develop a scalable genetic algorithm-based hyperparameter search for deep RL applications. HPS-RL packages its deep RL algorithms, optimization functions, and gym environments to allow researchers to explore deep RL benchmarks and also develop their algorithms. More importantly, using GA for multi-objective optimization, multiple hyperparameters are allowed to evolve over some generations to produce optimal robust deep RL solutions. We experimented with three gym environments, to show that over more generations, HPS-RL can find optimal RL architectures which can achieve higher rewards in fewer episodes. The main advantage of using GA over Bayesian Optimization is that they can be implemented in parallel while original Bayesian optimization process cannot. The only way of paralelizing  Bayesian optimization is that implementing acquisition function for multiple points \cite{Parallelacquisition}. Also, Bayesian approaches work well on continuous hyperparameters however in RL environments early episodes rely on randomness. Bayesian optimization requires large number of episodes to get optimized results. For early episodes, it behaves like random search. Moreover, the number of parameters is very important to get efficient result from Bayesian approach. As a conclusion, due to the random nature of RL problems in early stages, GA will provide more exploration and provide better parameters than that of Bayesian approach. 

Our future work will include extending HPS-RL with mpipy to allow distributed processing over large supercomputing to allow researchers to quickly find optimal solutions. Using GAs allows them to define the number of generations, which also allows researchers to work with a limited number of compute hours. HPS-RL has the potential to accelerate deep RL research. With a scalable search and easy-to-understand solution, researchers can automate the search and produce more reliable and robust deep RL architectures for RL applications.


\section{Acknowledgments}
This work was supported by the U.S. Department of Energy, Office of Science Early Career Research Program for `Large-scale Deep Learning for Intelligent Networks' Contract no  FP00006145.
\bibliographystyle{unsrt}  
\bibliography{example_paper}  

\end{document}